\documentclass[11pt,a4paper]{article}
\usepackage{amsmath,amssymb,amsthm,bbm}
\usepackage[dvipsnames]{xcolor}
\usepackage[normalem]{ulem}
\usepackage{tikz-qtree}
\usepackage{endnotes}
\usepackage{graphicx}
\usepackage{endnotes}
\usepackage{hyperref}
\usepackage{fancyhdr}
\usepackage{datetime}

\newcommand{\singbr}[1]{\left[#1\right]}

\newcommand{\paren}[1]{\left(#1\right)}

\newcommand{\expected}[2]{\underset{#1}{\E}\singbr{#2}}

\newcommand{\abs}[1]{\left|#1\right|}

\newcommand{\wh}[1]{\ensuremath{\widehat{#1}}}

\DeclareFontFamily{U}{mathx}{\hyphenchar\font45}
\DeclareFontShape{U}{mathx}{m}{n}{<-> mathx10}{}
\DeclareSymbolFont{mathx}{U}{mathx}{m}{n}
\DeclareMathAccent{\widebar}{0}{mathx}{"73}

\newcommand{\ep}{\ensuremath{\epsilon}}

\newcommand{\R}{\ensuremath{\mathbb{R}}}

\newcommand{\E}{\ensuremath{\textbf{E}}}

\newcommand{\ra}{\ensuremath{\rightarrow}}

\newcommand{\by}{\times}

\definecolor{light-gray}{gray}{0.80}

\definecolor{darkred}{rgb}{0.64, 0.0, 0.0}

\newtheorem{thm}{Theorem}[section]
\newtheorem*{thm*}{Theorem}

\DeclareMathOperator*{\argmax}{arg\,max}

\usepackage[hyperref]{naaclhlt2019}
\usepackage{times}
\usepackage{latexsym}
\usepackage{url}

\newcommand\kmeans{\textsc{$k$-means}}
\newcommand\glove{\textsc{g}lo\textsc{v}e}

\aclfinalcopy 
\def\aclpaperid{239} 

\title{Mutual Information Maximization for Simple and Accurate Part-Of-Speech Induction}

\author{Karl Stratos \\
  Toyota Technological Institute at Chicago \\
  {\tt stratos@ttic.edu}
}

\date{}

\begin{document}
\maketitle
\begin{abstract}
  We address part-of-speech (POS) induction by maximizing the mutual information between the induced label and its context.
  We focus on two training objectives that are amenable to stochastic gradient descent (SGD):
  a novel generalization of the classical Brown clustering objective and a recently proposed variational lower bound.
  While both objectives are subject to noise in gradient updates,
  we show through analysis and experiments that the variational lower bound is robust
  whereas the generalized Brown objective is vulnerable.
  We obtain strong performance on a multitude of datasets and languages
  with a simple architecture that encodes morphology and context.
\end{abstract}

\section{Introduction}

We consider information theoretic objectives for POS induction,
an important unsupervised learning problem in computational linguistics \cite{christodoulopoulos2010two}.
The idea is to make the induced label syntactically informative by maximizing its mutual information with respect to local context.
Mutual information has long been a workhorse in the development of NLP techniques,
for instance the classical Brown clustering algorithm \cite{brown1992class}.
But its role in today's deep learning paradigm is less clear and a subject of active investigation \citep{belghazi18a,oord2018representation}.

We focus on fully differentiable objectives that can be plugged into an automatic differentiation system and efficiently optimized by SGD.
Specifically, we investigate two training objectives.
The first is a novel generalization of the Brown clustering objective obtained by relaxing the hard clustering constraint.
The second is a recently proposed variational lower bound on mutual information \citep{mcallester2018information}.

A main challenge in optimizing these objectives is the difficulty of stochastic optimization.
Each objective involves entropy estimation which is a nonlinear function of all data and does not decompose over individual instances.
This makes the gradients estimated on minibatches inconsistent with the true gradient estimated from the entire dataset.
To our surprise, in practice we are able to optimize the variational objective effectively
but not the generalized Brown objective.
We analyze the estimated gradients and show that the inconsistency error is only logarithmic in the former but linear in the latter.

We validate our approach on POS induction by attaining strong performance on a multitude of datasets and languages.
Our simple architecture that encodes morphology and context reaches up to 80.1 many-to-one accuracy on the 45-tag Penn WSJ dataset
and achieves 4.7\% absolute improvement to the previous best result on the universal treebank. 
Unlike previous works, our model does not rely on computationally expensive structured inference or hand-crafted features.

\section{Background}

\subsection{Information Theory}

\paragraph{Mutual information.}
Mutual information between two random variables measures the amount of information gained about one variable by observing the other.
Unlike the Pearson correlation coefficient which only captures the degree of linear relationship,
mutual information captures any nonlinear statistical dependencies \citep{kinney2014equitability}.

Formally, the mutual information between discrete random variables $X, Y$ with a joint distribution $p$ is
the KL divergence between the joint distribution $p(x, y)$ and the product distribution $p(x)p(y)$ over $X, Y$:
\begin{align}
  I(X,Y)
  &= \sum_{x,y} p(x,y) \log \frac{p(x,y)}{p(x)p(y)} \notag \\
  &= \expected{(x, y)\sim p}{\log \frac{p(x,y)}{p(x)p(y)}} \label{eq:mi1}
\end{align}
It is thus nonnegative and zero iff $X$ and $Y$ are independent.
We assume that the marginals $p(x)$ and $p(y)$ are nonzero.

It is insightful to write mutual information in terms of entropy.
The entropy of $X$ is
\begin{align*}
  H(X) = - \sum_x p(x) \log p(x) = \expected{x\sim p}{\log \frac{1}{p(x)}}
\end{align*}
corresponding to the number of bits for encoding the behavior of $X$ under $p$.\footnote{The paper will always assume log base 2 to accommodate the bit interpretation.}
The entropy of $X$ given the information that $Y$ equals $y$ is
\begin{align*}
  H(X|Y=y) = - \sum_x p(x|y) \log p(x|y)
\end{align*}
Taking expectation over $Y$ yields the conditional entropy of $X$ given $Y$:
\begin{align}
  H(X|Y) 
  &= \sum_y p(y) \paren{ - \sum_x p(x|y) \log p(x|y) } \notag \\
  &= \expected{(x, y) \sim p}{\log \frac{1}{p(x|y)}} \label{eq:cent}
\end{align}
By manipulating the terms in mutual information, we can write
\begin{align*}
  I(X,Y)
  &= \expected{x \sim p}{\log \frac{1}{p(x)}}  - \expected{(x, y)\sim p}{\log \frac{1}{p(x|y)}} \\
  &= H(X) - H(X|Y)
\end{align*}
which expresses the amount of information on $X$ gained by observing $Y$.
Switching $X$ and $Y$ shows that $I(X,Y) = H(Y) - H(Y|X)$.

\paragraph{Cross entropy.}
If $p$ and $q$ are full-support distributions over the same discrete set,
the cross entropy between $p$ and $q$ is the asymmetric quantity
\begin{align*}
  H(p,q) = - \sum_x p(x) \log q(x) = \expected{x \sim p}{\log \frac{1}{q(x)}}
\end{align*}
corresponding to the number of bits for encoding the behavior of $X$ under $p$ by using $q$.
It is revealing to write entropy in terms of KL divergence. Multiplying the term inside the log by $p(x)/p(x)$ we derive
\begin{align*}
  H(p,q)
  &= \expected{x \sim p}{\log \frac{1}{p(x)}} + \expected{x \sim p}{\log \frac{p(x)}{q(x)}} \\
  &= H(p) + D_{\mathrm{KL}}(p||q)
\end{align*}
Thus $H(p, q) \geq H(p)$ with equality iff $p = q$.

\subsection{Brown Clustering}

Our primary inspiration comes from Brown clustering \citep{brown1992class}, a celebrated word clustering technique
that had been greatly influential in unsupervised and semi-supervised NLP long before
continuous representations based on neural networks were popularized.
It finds a clustering $C:V \rightarrow [m]$ of the vocabulary $V$
into $m$ classes by optimizing the mutual information between the clusters of a random bigram $(X, Y)$.
Given a corpus of $N$ words $(x_1 \ldots x_N)$, it assumes a uniform distribution over consecutive word pairs $(x_{i-1}, x_i)$ and optimizes the following empirical objective
\begin{align}
  \max_{C:V \rightarrow [m]} \sum_{c, c' \in [m]} \frac{\#(c, c')}{N}  \log \paren{ \frac{\#(c, c')  N}{\#(c)  \#(c') }} \label{eq:brown}
\end{align}
where $\#(c, c')$ denotes the number of occurrences of the cluster pair $(c, c')$ under $C$.
While this optimization is intractable, \citet{brown1992class} derive an effective heuristic
that 1.~initializes $m$ most frequent words as singleton clusters and 2.~repeatedly merges
a pair of clusters that yields the smallest decrease in mutual information.
The resulting clusters have been useful in many applications \citep{koo2008simple,owoputi2013improved}
and has remained a strong baseline for POS induction decades later \cite{christodoulopoulos2010two}.
But the approach is tied to highly nontrivial combinatorial optimization tailored for the specific problem
and difficult to scale/generalize.

\section{Objectives}
\label{sec:objectives}

In the remainder of the paper, we assume discrete random variables $(X, Y) \in \mathcal{X} \by \mathcal{Y}$ with a joint distribution $D$ that represent naturally co-occurring observations.
In POS induction experiments, we will set $D$ to be a context-word distribution where $Y$ is a random word and $X$ is the surrounding context of $Y$ (thus $\mathcal{Y}$ is the vocabulary and $\mathcal{X}$ is the space of all possible contexts).
Let $m$ be the number of labels to induce. We introduce a pair of trainable classifiers
that define conditional label distributions $p(z|x)$ and $q(z|y)$ for all $x \in \mathcal{X}$, $y \in \mathcal{Y}$, and $z \in [m]$.
For instance, $p(\cdot|y)$ can be the output of a softmax layer on some transformation of $y$.

Our goal is to learn these classifiers without observing the latent variable $z$ by optimizing an appropriate objective.
For training data, we assume $N$ iid samples $(x_1, y_1) \ldots (x_N, y_N) \sim D$.

\subsection{Generalized Brown Objective}
\label{sec:gbrown}

Our first attempt is to maximize the mutual information between the predictions of $p$ and $q$.
Intuitively, this encourages $p$ and $q$ to agree on some annotation scheme (up to a permutation of labels),
modeling the dynamics of inter-annotator agreement \cite{artstein2017inter}.
It can be seen as a differentiable generalization of the Brown clustering objective.
To this end, define
\begin{align*}
  p(z) &= \expected{x \sim D}{ p(z|x) } &&&\forall z \in [m]\\
  q(z) &= \expected{y \sim D}{ q(z|y) } &&&\forall z \in [m]
\end{align*}
The mutual information between the predictions of $p$ and $q$ on a single sample $(x, y)$ is then
\begin{align*}
  J^{\mathrm{mi}}_{x,y} = \sum_{z, z'} p(z|x) q(z'|y) \log \frac{p(z|x) q(z'|y)}{p(z)q(z')}
\end{align*}
and the objective (to maximize) is
\begin{align*}
  J^{\mathrm{mi}} =  \expected{(x, y) \sim D}{ J^{\mathrm{mi}}_{x,y} }
\end{align*}
Note that this becomes exactly the original Brown objective \eqref{eq:brown} if $(X, Y)$ is defined as a random bigram and $p$ and $q$ are tied and constrained to be a hard clustering.

\paragraph{Empirical objective.}
In practice, we work with the value of empirical mutual information $\wh{J}^{\mathrm{mi}}$ estimated from the training data:
\begin{align}
  \hat{p}(z) &= \frac{1}{N} \sum_{i=1}^N  p(z|x_i) \hspace{20mm} \forall z \in [m]\notag  \\
  \hat{q}(z) &= \frac{1}{N} \sum_{i=1}^N  q(z|y_i) \hspace{21mm} \forall z \in [m] \notag \\
  \wh{J}^{\mathrm{mi}}_i &= \sum_{z, z'} p(z|x_i) q(z'|y_i) \log \frac{p(z|x_i) q(z'|y_i)}{\hat{p}(z)\hat{q}(z')} \notag\\
  \wh{J}^{\mathrm{mi}} &=  \frac{1}{N} \sum_{i=1}^N \wh{J}^{\mathrm{mi}}_i \label{eq:mi-obj}
\end{align}
Our task is to maximize \eqref{eq:mi-obj} by taking gradient steps at random minibatches.
Note, however, that the objective cannot be written as a sum of local objectives
because we take log of the estimates $\hat{q}(z)$ and $\hat{p}(z)$ computed from all samples.
This makes the stochastic gradient estimator biased (i.e., it does not match the gradient of \eqref{eq:mi-obj} in expectation)
and compromises the correctness of SGD.
This bias is investigated more closely in Section~\ref{sec:analysis}.

\subsection{Variational Lower Bound}

The second training objective we consider can be derived in a rather heuristic but helpful manner as follows.
Since $X,Y$ are always drawn together, if $q(z|y)$ is the target label distribution for the pair $(x, y)$,
then we can train $p(z|x)$ by minimizing the cross entropy between $q$ and $p$ over samples
\begin{align*}
  H(q,p) &=  \expected{(x,y)\sim D}{ - \sum_{z} q(z|y) \log p(z|x)}
\end{align*}
which is minimized to zero at $p = q$.
However, $q$ is also untrained and needs to be trained along with $p$.
Thus this loss alone admits trivial solutions such as setting $p(1|x) = p(1|y) = 1$ for all $(x, y)$.
This undesirable behavior can be prevented by simultaneously maximizing the entropy of $q$.
Let $Z$ denote a random label from $q$ with distribution $q(z) = \expected{y \sim D}{ q(z|y) } $ (thus $Z$ is a function of $q$).
The entropy of $Z$ is
\begin{align*}
  H(Z) &=  - \sum_{z} q(z) \log q(z)
\end{align*}
Putting together, the objective (to maximize) is
\begin{align*}
 J^{\mathrm{var}} = H(Z) - H(q, p) 
\end{align*}

\paragraph{Variational interpretation.}
The reason this objective is named a variational lower bound is due to \citet{mcallester2018information}
who shows the following.
Consider the mutual information between $Z$  and the raw signal $X$:
\begin{align}
  I(X,Z) = H(Z) - H(Z|X) \label{eq:david}
\end{align}
Because $Z$ is drawn from $q$ conditioning on $Y$, which is co-distributed with $X$, we have a Markov chain
$X \ra Y \xrightarrow{q} Z$.
Thus maximizing $I(X,Z)$ over the choice of $q$ is a reasonable objective that enforces ``predictive coding'':
the label predicted by $q$ from $Y$ must be as informative of $X$ as possible.
It can be seen as a special case of the objective underlying the information bottleneck method \citep{tishby2000information}.

So what is the problem with optimizing \eqref{eq:david} directly?
The problem is that the conditional entropy under the model
\begin{align}
  H(Z|X) =  \expected{\substack{(x,y)\sim D\\ z \sim q(\cdot|y)}}{  \log \frac{1}{\pi(z|x)}} \label{eq:hz_x}
\end{align}
involves the posterior probability of $z$ given $x$
\begin{align*}
  \pi\paren{ z | x} = \frac{\sum_y D(x, y) q(z|y)} {\sum_{y,z} D(x, y) q(z|y)}
\end{align*}
This \emph{conditional} marginalization is generally intractable and cannot be approximated by sampling
since the chance of seeing a particular $x$ is small.
However, we can introduce a variational distribution $p(z|x)$ to model $\pi\paren{ z | x}$.
Plugging this into \eqref{eq:hz_x} we observe that
\begin{align*}
  &\expected{\substack{(x,y)\sim D\\ z \sim q(\cdot|y)}}{  \log \frac{1}{p(z|x)}} = H(q, p)
\end{align*}
Moreover,
\begin{align*}
  &\expected{\substack{(x,y)\sim D\\ z \sim q(\cdot|y)}}{  \log \frac{1}{p(z|x)}} \\
  &\hspace{10mm}= \expected{\substack{(x,y)\sim D\\ z \sim q(\cdot|y)}}{  \log \frac{\pi\paren{ z | x}}{\pi\paren{ z | x}p(z|x)}} \\
  &\hspace{10mm}= H(Z|X) + D_{\mathrm{KL}}(\pi||p)
\end{align*}
Thus $H(q, p)$ is an upper bound on $H(Z|X)$ for any $p$ with equality iff $p$ matches the true posterior distribution $\pi$. 
This in turn means that $J^{\mathrm{var}} = H(Z) - H(q, p)$ is a lower bound on $I(X,Z)$, hence the name.

\paragraph{Empirical objective.}

As with the generalized Brown objective, the variational lower bound can be estimated from the training data as
\begin{align}
  \wh{H}(Z) &= - \sum_{z} \hat{q}(z) \log \hat{q}(z) \notag \\
  \wh{H}(q,p) &=  \frac{1}{N} \sum_{i=1}^N \paren{ - \sum_{z} q(z|y_i) \log p(z|x_i) } \notag\\
  \wh{J}^{\mathrm{var}} &= \wh{H}(Z) - \wh{H}(q,p) \label{eq:var-obj}
\end{align}
where $\hat{q}(z)$ is defined as in Section~\ref{sec:gbrown}.
Our task is again to maximize this empirical objective \eqref{eq:var-obj} by taking gradient steps at random minibatches.
Once again, it cannot be written as a sum of local objectives
because the entropy term involves log of $\hat{q}(z)$ computed from all samples.
Thus it is not clear if stochastic optimization will be effective.

\section{Analysis}
\label{sec:analysis}

The discussion of the two objectives in the previous section
is incomplete because the stochastic gradient estimator is biased under both objectives.
In this section, we formalize this issue and analyze the bias.

\subsection{Setting}
\label{sec:setting}

Let $B_1 \ldots B_K$ be a partition of the $N$ training examples $(x_1, y_1) \ldots (x_N, y_N)$ into $K$ (iid) minibatches.
For simplicity, assume $\abs{B_k} = M$ for all $k$ and $N = MK$.
We will adopt the same notation in Section~\ref{sec:objectives} for $\hat{p}(z)$ and $\hat{q}(z)$ estimated from all $N$ samples.
Define analogous estimates based on the $k$-th minibatch by
\begin{align*}
  \hat{p}_k(z) &= \frac{1}{M} \sum_{x \in B_k}  p(z|x) &&\forall z \in [m] \\
  \hat{q}_k(z) &= \frac{1}{M} \sum_{y \in B_k}  q(z|y) &&\forall z \in [m]
\end{align*}
If $l_N$ denotes an objective function computed from all $N$ samples in the training data
and $l_k$ denotes the same objective computed from $B_k$,
a condition on the correctness of SGD is that the gradient of $l_k$ (with respect to model parameters)
is consistent with the gradient of $l_N$ on average:
\begin{align}
  \nabla l_N = \frac{1}{K} \sum_{k=1}^K \nabla l_k + \ep \label{eq:sgd}
\end{align}
where $\ep$ denotes the bias of the stochastic gradient estimator.
In particular, any loss of the form $l_N = (1/K) \sum_k l_k $
that decomposes over independent minibatches (e.g., the cross-entropy loss for supervised classification)  satisfies \eqref{eq:sgd} with $\ep = 0$.
The bias is nonzero for the unsupervised objectives considered in this work due to the issues discussed in Section~\ref{sec:objectives}.

\subsection{Result}

The following theorem precisely quantifies the bias for the empirical losses associated with the variational bound and the generalized Brown objectives.
We only show the result with the gradient with respect to $q$, but the result with the gradient with respect to $p$ is analogous and omitted for brevity.

\begin{thm}
  Assume the setting in Section~\ref{sec:setting} and
  the gradient is taken with respect to the parameters of $q$.
  For $l_N = - \wh{J}^{\mathrm{var}}$ defined in \eqref{eq:var-obj},
  \begin{align*}
    \ep &= \frac{1}{K} \sum_{k=1}^K \sum_z  \log \frac{\hat{q}(z)}{\hat{q}_k(z)} \nabla \hat{q}_k(z)
  \end{align*}
  On the other hand, for $l_N = - \wh{J}^{\mathrm{mi}}$ defined in \eqref{eq:mi-obj},
  \begin{align*}
    \ep &= \frac{1}{N} \sum_{k=1}^K \sum_{z, z'} \bigg( \ep_k(z,z') \nabla \hat{q}_k(z') \\
    &\hspace{5mm}+ \log \frac{\hat{p}(z)\hat{q}(z')}{\hat{p}_k(z)\hat{q}_k(z')}  \sum_{(x,y) \in B_k}  p(z|x) \nabla q(z'|y)\bigg)
  \end{align*}
  where
  \begin{align*}
    \ep_k(z,z') &= \frac{1}{K} \sum_{i=1}^N \frac{p(z|x_i) q(z'|y_i)}{\hat{q}(z')}   \\
    &\hspace{5mm}- \sum_{(x, y) \in B_k} \frac{p(z|x) q(z'|y)}{\hat{q}_k(z')}
  \end{align*}
  \label{thm:1}
\end{thm}

A proof can be found in the appendix.
We see that both biases go to zero as $\hat{p}_k$ and $\hat{q}_k$ approach $\hat{p}$ and $\hat{q}$. 
However, the bias is logarithmic in the ratio $\hat{q}(z)/\hat{q}_k(z)$ for the variational lower bound
but roughly linear in the difference between $\frac{1}{\hat{q}(z')}$ and $\frac{1}{\hat{q}_k(z') }$ for the generalized Brown objective.
In this sense, the variational lower bound is exponentially more robust to noise in minibatch estimates
than the generalized Brown objective.
This is confirmed in experiments: we are able to optimize $\wh{J}^{\mathrm{var}}$ with minibatches as small as $80$ examples
but unable to optimize $\wh{J}^{\mathrm{mi}}$ unless minibatches are prohibitively large.

\section{Experiments}
\label{sec:experiments}

We demonstrate the effectiveness of our training objectives on the task of POS induction.
The goal of this task is to induce the correct POS tag for a given word in context \cite{merialdo1994tagging}.
As typical in unsupervised tasks, evaluating the quality of induced labels is challenging; see \citet{christodoulopoulos2010two} for an in-depth discussion.
To avoid complications, we follow a standard practice \cite{berg2010painless,ammar2014conditional,lin-EtAl:2015:NAACL-HLT,TACL837}
and adopt the following setting for all compared methods.
\begin{itemize}
\item We use many-to-one accuracy as a primary evaluation metric.
  That is, we map each induced label to the most frequently coinciding ground-truth POS tag
  in the annotated data and report the resulting accuracy.
  We also use the V-measure \citep{rosenberg2007v} when comparing with CRF autoencoders to be consistent with reported results \cite{ammar2014conditional,lin-EtAl:2015:NAACL-HLT}.
\item We use the number of ground-truth POS tags as the value of $m$ (i.e., number of labels to induce).
  This is a data-dependent quantity, for instance 45 in the Penn WSJ and 12 in the universal treebank.
  Fixing the number of tags this way obviates many evaluation issues.
\item Model-specific hyperparameters are tuned on the English Penn WSJ dataset.
  This configuration is then fixed and used for all other datasets: 10 languages in the universal treebank\footnote{https://github.com/ryanmcd/uni-dep-tb}
  and 7 languages from CoNLL-X and CoNLL 2007.
\end{itemize}

\subsection{Setting}
We set $D$ to be a uniform distribution over context-word pairs in the training corpus.
Given $N$ samples $(x_1, y_1) \ldots (x_N, y_N) \sim D$, we optimize the variational objective \eqref{eq:var-obj} or the generalized Brown objective \eqref{eq:mi-obj}
by taking gradient steps at random minibatches.
This gives us conditional label distributions $p(z|x)$ and $q(z|y)$ for all contexts $x$, words $y$, and labels $z$.
At test time, we use
\begin{align*}
  z^* = \argmax_z q(z|y)
\end{align*}
as the induced label of word $y$. We experimented with different inference methods such as taking $\argmax_z p(z|x)q(z|y)$ but did not find it helpful.

\subsection{Definition of $(X,Y)$}
Let $V$ denote the vocabulary.
We assume an integer $H \geq 1$ that specifies the width of local context.
Given random word $y \in V$, we set $x \in V^{2H}$ to be an ordered list of $H$ left and $H$ right words of $y$.
For example, with $H = 2$, a typical context-target pair $(x, y) \sim D$ may look like
\begin{align*}
  x &= (\mbox{``had'', ``these'', ``in'', ``my''}) \\
  y &= \mbox{``keys''}
\end{align*}
We find this simple fixed-window definition of observed variables to be the best inductive bias for POS induction.
The correct label can be inferred from either $x$ or $y$ in many cases: in the above example,
we can infer that the correct POS tag is plural noun by looking at the target or the context.

\subsection{Architecture}

We use the following simple architecture to parameterize the label distribution $p(\cdot|x)$ conditioned on context $x \in V^H$ and
the label distribution $q(\cdot|y)$ conditioned on word $y \in V$.

\paragraph{Context architecture.}
The parameters of $p$ are word embeddings $e_w \in \R^d$ for all $w \in V$
and matrices $W_j \in \R^{m \by d}$ for all $j = 1 \ldots 2H$.
Given $2H$ ordered contextual words $x = (w_j)_{j=1}^{2H}$,
we define
\begin{align*}
  p\paren{\cdot | x} &= \mbox{softmax} \paren{\sum_{j = 1}^{2H} W_j e_{w_j}}
\end{align*}

\begin{figure}[t!]
  \begin{center}
    \includegraphics[scale=0.8]{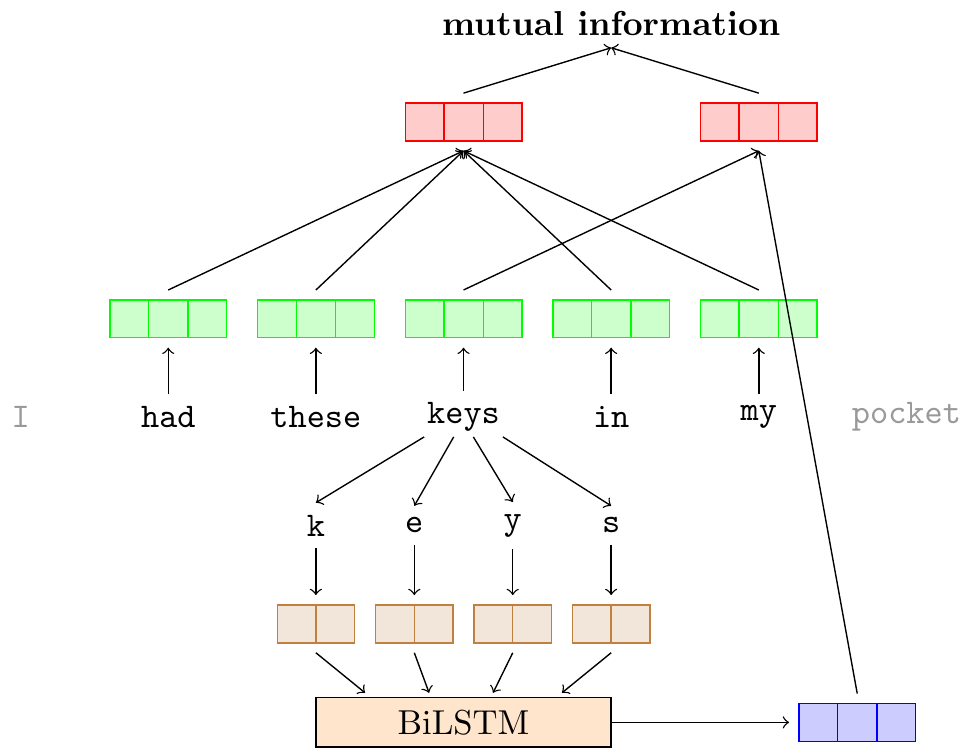}
    \caption{Architecture illustrated on the example text ``had these keys in my'' with target $Y = \mbox{``keys''}$.}
    \label{fig:arch}
  \end{center}
\end{figure}

\paragraph{Word architecture.}
The parameters of $q$ are the same word embeddings $e_w \in \R^d$ shared with $p$,
character embeddings $e_c \in \R^{d/2}$ for all distinct characters $c$,
two single-layer LSTMs with input/output dimension $d/2$,
and matrices $W_c, W_w \in \R^{m \by d}$.
Given the word $y$ with character sequence $c_1 \ldots c_T$,
we define
\begin{align*}
  (f_1 \ldots f_T) &= \mbox{LSTM}_f(e_{c_1} \ldots e_{c_T}) \\
  (b_1 \ldots b_T) &= \mbox{LSTM}_b(e_{c_T} \ldots e_{c_1}) \\
  q(\cdot | y) &= \mbox{softmax} \paren{W_c \begin{bmatrix}f_T \\b_T \end{bmatrix} + W_w e_y}
\end{align*}
The overall architecture is illustrated in Figure~\ref{fig:arch}.
Our hyperparameters are the embedding dimension $d$, the context width $H$, the learning rate, and the minibatch size.\footnote{An implementation is available at: \url{https://github.com/karlstratos/mmi-tagger}.}
They are tuned on the 45-tag Penn WSJ dataset to maximize accuracy.
The final hyperparameter values are given in the table:
\begin{center}
  \begin{tabular}{|c|c|}
    \hline
    $d$ & $200$ \\
    \hline
    $H$ & $2$ \\
    \hline
    learning rate & 0.001 \\
    \hline
    minibatch size & 80 \\
    \hline
  \end{tabular}
\end{center}

\subsection{Baselines}

We focus on comparing with the following models which are some of the strongest baselines in the literature we are aware of.
\citet{berg2010painless} extend a standard hidden Markov Model (HMM) to incorporate linguistic features.
\citet{TACL837} develop a factorization-based algorithm for learning a constrained HMM.
\citet{ammar2014conditional} propose a CRF autoencoder that reconstructs words from a structured label sequence.
\citet{lin-EtAl:2015:NAACL-HLT} extend \citet{ammar2014conditional} by switching a categorical reconstruction distribution with a Gaussian distribution.
In addition to these baselines, we also report results with Brown clustering \citep{brown1992class},
the Baum-Welch algorithm \citep{baum1966statistical},
and $k$-means clustering of $300$-dimensional GloVe vectors \cite{pennington2014glove}.

\subsection{Results}

\paragraph{The 45-tag Penn WSJ dataset.}
The 45-tag Penn WSJ dataset is a corpus of around one million words each tagged with one of $m = 45$ tags.
It is used to optimize hyperparameter values for all compared methods.
Table~\ref{tab:en45} shows the average accuracy over $10$ random restarts with the best hyperparameter configurations;
standard deviation is given in parentheses (except for deterministic methods \citet{TACL837} and Brown clustering).

Our model trained with the variational objective \eqref{eq:var-obj} outperforms all baselines.\footnote{
  We remark that \citet{tran2016unsupervised} report a single number 79.1 with a neuralized HMM.
  We also note that the concurrent work by \citet{he2018unsupervised} obtains 80.8 by using word embeddings carefully pretrained on one billion words.
}
We also observe that our model trained with the generalized Brown objective \eqref{eq:mi-obj} does not work.
We have found that unless the minibatch size is as large as 10,000 the gradient steps do not effectively increase the
true data-wide mutual information \eqref{eq:mi-obj}.
This supports our bias analysis in Section~\ref{sec:analysis}.
While it may be possible to develop techniques to resolve the difficulty, for instance keeping a moving average of estimates
to stabilize estimation, we leave this as future work and focus on the variational objective in the remainder of the paper.

\begin{table}[t!]
  \begin{center}
    {
      \begin{tabular}{|l||l|}
        \hline
        Method &  Accuracy    \\

        \hline
        Variational $\wh{J}^{\mathrm{var}}$ \eqref{eq:var-obj}             & $\boldsymbol{78.1} \;\mbox{\tiny (${\pm 0.8})$}$ \\
        Generalized Brown $\wh{J}^{\mathrm{mi}}$ \eqref{eq:mi-obj}          & $48.8 \;\;\mbox{\tiny (${\pm 0.9})$}$ \\
        \hline
            \citet{berg2010painless}
                   &  $74.9 \;\;\mbox{\scriptsize (${\pm 1.5})$}$ \\
            \citet{TACL837}
                &  $67.7$  \\
            \citet{brown1992class}
                 &  $65.6$  \\
            Baum-Welch
                     &  $62.6 \;\;\mbox{\scriptsize (${\pm 1.1})$}$ \\
            \kmeans{}     &  $32.6 \;\;\mbox{\tiny (${\pm 0.7})$}$ \\
        \hline
      \end{tabular}

      \caption{Many-to-one accuracy on the 45-tag Penn WSJ with the
        best hyperparameter configurations.
        The average accuracy over $10$ random restarts is reported
        and the standard deviation is given in parentheses (except for deterministic methods).
      }
      \label{tab:en45}
    }
  \end{center}
\end{table}

Table~\ref{tab:ablation-experiments} shows ablation experiments on our best model (accuracy 80.1) to better understand the sources of its strong performance.
Context size $H=2$ is a sizable improvement over $H=3$ or $H=1$.
Random sampling is significantly more effective than sentence-level batching (i.e., each minibatch is the set of context-word pairs within a single sentence as done in \citet{mcallester2018information}).
\glove{} initialization of word embeddings $e_w$ is harmful.
As expected for POS tagging, morphological modeling with LSTMs gives the largest improvement.

While it may be surprising that \glove{} initialization is harmful, it is well known that pretrained word embeddings
do not necessarily capture syntactic relationships (as evident in the poor performance of $k$-means clustering).
Consider the top ten nearest neighbors of the word ``made'' under GloVe embeddings (840B.300d, within PTB vocab):
\begin{center}
  {\small
  \begin{tabular}{|c|c|}
    \hline
    Cosine Similarity & Nearest Neighbor \\
    \hline
    0.7426&		making \\
    0.7113&		make \\
    0.6851&		that \\
    0.6613&		they \\
    0.6584&		been \\
    0.6574&		would \\
    0.6533&		brought \\
    0.6521&		had \\
    0.6514&		came \\
    0.6494&		but \\
    0.6486&		even \\
    \hline
  \end{tabular}
  }
\end{center}
The neighbors are clearly not in the same syntactic category. The embeddings can be made more syntactic by controlling the context window.
But we found it much more effective (and much simpler) to start from randomly initialized embeddings and let the objective induce appropriate representations.

\begin{table}[t!]
  \begin{center}
    {
      \begin{tabular}{|c|c|}
        \hline
        Configuration                 & Accuracy \\
        \hline
        Best                                   & 80.1   \\
        $H = 3$                                & 75.9   \\
        $H = 1$                                & 75.9   \\
        Sentence-level batching                & 72.4   \\
        \glove{} initialization                & 67.6   \\
        No character encoding                  & 65.6   \\
        \hline
      \end{tabular}
    }
    \caption{Ablation of the best model on Penn WSJ.}
    \label{tab:ablation-experiments}
  \end{center}
\end{table}

\paragraph{The 12-tag universal treebank.}
The universal treebank v2.0 is a corpus in ten languages tagged with $m=12$ universal POS tags \cite{mcdonald2013universal}.
We use this corpus to be compatible with existing results.
Table~\ref{tab:univ12} shows results on the dataset, using the same setting in the experiments on the Penn WSJ dataset.
Our model significantly outperforms the previous state-of-the-art, achieving an absolute gain of 4.7 over \citet{TACL837} in average accuracy.

\begin{table*}[t!]
{\small
\begin{center}
  \begin{tabular}{|c||c|c|c|c|c|c|c|c|c|c||c|}
    \hline
    Method                                     & de & en & es & fr & id & it & ja & ko & pt-br & sv & Mean\\
    \hline
Variational $\wh{J}^{\mathrm{var}}$ \eqref{eq:var-obj} & \begin{tabular}{@{}c@{}}{\scriptsize($\pm$1.5)} \\ \textbf{75.4} \end{tabular} & \begin{tabular}{@{}c@{}}{\scriptsize($\pm$1.7)} \\ \textbf{73.1} \end{tabular} & \begin{tabular}{@{}c@{}} {\scriptsize($\pm$1.0)} \\ 73.1\end{tabular} & \begin{tabular}{@{}c@{}} {\scriptsize($\pm$2.9)} \\ 70.4\end{tabular} & \begin{tabular}{@{}c@{}} {\scriptsize($\pm$1.5)} \\ \textbf{73.6}\end{tabular} & \begin{tabular}{@{}c@{}} {\scriptsize($\pm$3.3)} \\ \textbf{67.4}\end{tabular} & \begin{tabular}{@{}c@{}} {\scriptsize($\pm$0.4)} \\ 77.9\end{tabular} & \begin{tabular}{@{}c@{}}{\scriptsize($\pm$1.2)} \\ \textbf{65.6}\end{tabular} & \begin{tabular}{@{}c@{}}{\scriptsize($\pm$2.3)} \\ \textbf{70.7}\end{tabular} & \begin{tabular}{@{}c@{}}{\scriptsize($\pm$1.5)} \\ \textbf{67.1} \end{tabular} & \begin{tabular}{@{}c@{}}{} \\ \textbf{71.4} \end{tabular} \\
      \hline
                        \citeauthor{TACL837}                   & 63.4          & 71.4          & \textbf{74.3}          & \textbf{71.9}          & 67.3          & 60.2         & 69.4         & 61.8            & 65.8          & 61.0  & 66.7\\
            \citeauthor{berg2010painless} & \begin{tabular}{@{}c@{}}{\scriptsize($\pm$1.8)} \\ 67.5 \end{tabular} & \begin{tabular}{@{}c@{}}{\scriptsize($\pm$3.5)} \\ 62.4 \end{tabular} & \begin{tabular}{@{}c@{}} {\scriptsize($\pm$3.1)} \\ 67.1\end{tabular} & \begin{tabular}{@{}c@{}} {\scriptsize($\pm$4.5)} \\ 62.1\end{tabular} & \begin{tabular}{@{}c@{}} {\scriptsize($\pm$3.9)} \\ 61.3\end{tabular} & \begin{tabular}{@{}c@{}} {\scriptsize($\pm$2.9)} \\ 52.9\end{tabular} & \begin{tabular}{@{}c@{}} {\scriptsize($\pm$2.9)} \\ \textbf{78.2}\end{tabular} & \begin{tabular}{@{}c@{}}{\scriptsize($\pm$3.6)} \\ 60.5\end{tabular} & \begin{tabular}{@{}c@{}}{\scriptsize($\pm$2.2)} \\ 63.2\end{tabular} & \begin{tabular}{@{}c@{}}{\scriptsize($\pm$2.5)} \\ 56.7 \end{tabular} & \begin{tabular}{@{}c@{}}{} \\ 63.2 \end{tabular}\\
              \citeauthor{brown1992class}                    & 60.0          & 62.9          & 67.4          & 66.4          & 59.3          & 66.1         & 60.3         & 47.5            & 67.4          & 61.9  & 61.9\\
Baum-Welch          & \begin{tabular}{@{}c@{}}{\scriptsize($\pm$4.8)} \\ 45.5 \end{tabular} & \begin{tabular}{@{}c@{}}{\scriptsize($\pm$3.4)} \\ 59.8 \end{tabular} & \begin{tabular}{@{}c@{}} {\scriptsize($\pm$2.2)} \\ 60.6\end{tabular} & \begin{tabular}{@{}c@{}} {\scriptsize($\pm$3.6)} \\ 60.1\end{tabular} & \begin{tabular}{@{}c@{}} {\scriptsize($\pm$3.1)} \\ 49.6\end{tabular} & \begin{tabular}{@{}c@{}} {\scriptsize($\pm$2.6)} \\ 51.5\end{tabular} & \begin{tabular}{@{}c@{}} {\scriptsize($\pm$2.1)} \\ 59.5\end{tabular} & \begin{tabular}{@{}c@{}}{\scriptsize($\pm$0.6)} \\ 51.7\end{tabular} & \begin{tabular}{@{}c@{}}{\scriptsize($\pm$3.7)} \\ 59.5\end{tabular} & \begin{tabular}{@{}c@{}}{\scriptsize($\pm$3.0)} \\ 42.4 \end{tabular}  & \begin{tabular}{@{}c@{}}{} \\ 54.0 \end{tabular}\\
              \hline
\end{tabular}
  \caption{
    Many-to-one accuracy on the 12-tag universal treebank dataset. We use the same setting in Table~\ref{tab:en45}.
    All models use a fixed hyperparameter configuration optimized on the 45-tag Penn WSJ.
  }
\label{tab:univ12}
\end{center}
}
\end{table*}

\paragraph{Comparison with CRF autoencoders.}
Table~\ref{tab:ammar} shows a direct comparison with CRF autoencoders \citep{ammar2014conditional,lin-EtAl:2015:NAACL-HLT} in many-to-one accuracy and the V-measure.
We compare against their reported numbers by running our model once on the same datasets using the same setting in the experiments on the Penn WSJ dataset.
The data consists of the training portion of CoNLL-X and CoNLL 2007 labeled with 12 universal tags.
Our model is competitive with all baselines.

\begin{table*}[t!]
  {\small
    \begin{center}
      \begin{tabular}{|c||c||c|c|c|c|c|c|c||c|}
        \hline
        Metric & Method    &  Arabic       & Basque            & Danish        & Greek         & Hungarian       & Italian       & Turkish        & Mean       \\
        \hline
        M2O    & Variational $\wh{J}^{\mathrm{var}}$ \eqref{eq:var-obj}  & \textbf{74.3} & \textbf{70.4} & \textbf{71.7} & \textbf{66.1} & \textbf{61.2} & \textbf{67.4} & \textbf{64.2} & \textbf{67.9}   \\
        \cline{2-10}
        & \citeauthor{ammar2014conditional}  & 69.1 &68.1 &60.9 &63.5 &57.1 &60.4 &60.4 &62.8 \\
        & \citeauthor{berg2010painless}  & 66.8 &66.2& 60.0 &60.2 &56.8	& 64.1&62.0& 62.3   \\
        & Baum-Welch     & 49.7	         &  44.9         & 42.4          & 39.2	         & 45.2          & 39.3 & 52.7 & 44.7 \\
        \hline
        \hline
        VM     & Variational $\wh{J}^{\mathrm{var}}$ \eqref{eq:var-obj}  & \textbf{56.9} & 43.6              & \textbf{56.0} & \textbf{56.3} & 47.9            & \textbf{53.3} & 38.5           & \textbf{50.4}   \\
        \cline{2-10}
        & \citeauthor{lin-EtAl:2015:NAACL-HLT}      & 50.5          & \textbf{51.7}     &  51.3         & 50.0          & \textbf{55.9}   & 46.3          & \textbf{43.1}  & 49.8             \\
        & \citeauthor{ammar2014conditional}         & 49.1&	41.1&	46.1&	49.1&	41.1&	43.1&	35.0&	43.5 \\
        & \citeauthor{berg2010painless}   & 33.8&	33.4&	41.1&	40.9&	39.0&	46.6&	31.6&	38.8 \\
        & Baum-Welch     & 15.3&	8.2&	11.1&	9.6&	10.1&	9.9&	11.6&	10.8 \\
        \hline
      \end{tabular}
      \caption{
        Comparison with the reported results with CRF autoencoders in many-to-one accuracy (M2O) and the V-measure (VM).
      }
      \label{tab:ammar}
    \end{center}
}
\end{table*}

\section{Related Work}

Information theory, in particular mutual information, has played a prominent role in NLP \citep{church1990word,brown1992class}.
It has intimate connections to the representation learning capabilities of neural networks \citep{tishby2015deep}
and underlies many celebrated modern approaches to unsupervised learning such as generative adversarial networks (GANs) \citep{goodfellow2014generative}.

There is a recent burst of effort in learning continuous representations by optimizing various lower bounds on mutual information
\citep{belghazi18a,oord2018representation,hjelm2018learning}.
These representations are typically evaluated on extrinsic tasks as features.
In contrast, we learn discrete representations by optimizing a novel generalization of
the Brown clustering objective \citep{brown1992class} and a variational lower bound on mutual information proposed by \citet{mcallester2018information}.
We focus on intrinsic evaluation of these representations on POS induction.
Extrinsic evaluation of these representations in downstream tasks is an important future direction.

The issue of biased stochastic gradient estimators is a common challenge in unsupervised learning (e.g., see \citealp{wang2015stochastic}).
This arises mainly because the objective involves a nonlinear transformation of all samples in a training dataset,
for instance the whitening constraints in deep canonical correlation analysis (CCA) \citep{andrew2013deep}.
In this work, the problem arises because of entropy.
This issue is not considered in the original work of \citet{mcallester2018information} and the error analysis we present in Section~\ref{sec:analysis} is novel.
Our finding is that the feasibility of stochastic optimization greatly depends on the size of the bias in gradient estimates,
as we are able to effectively optimize the variational objective while not the generalized Brown objective.

Our POS induction system has some practical advantages over previous approaches.
Many rely on computationally expensive structured inference or pre-optimized features (or both).
For instance, \citet{tran2016unsupervised} need to calculate forward/backward messages and is limited to truncated sequences by memory constraints.
\citet{berg2010painless} rely on extensively hand-engineered linguistic features.
\citet{ammar2014conditional}, \citet{lin-EtAl:2015:NAACL-HLT}, and \citet{he2018unsupervised} rely on carefully pretrained lexical representations like Brown clusters and word embeddings.
In contrast, the model presented in this work requires no expensive structured computation or feature engineering and uses word/character embeddings trained from scratch.
It is easy to implement using a standard neural network library and outperforms these previous works in many cases.

\section*{Acknowledgments}

The author thanks David McAllester for many insightful discussions,
and Sam Wiseman for helpful comments.
The Titan Xp used for this research was donated by the NVIDIA Corporation.

\bibliography{naaclhlt2019}
\bibliographystyle{acl_natbib}

\appendix

\section{Proof of Theorem~4.1}

We first analyze the variational loss
\begin{align*}
 \wh{H}(q,p) - \wh{H}(Z)
\end{align*}
Note that the cross entropy term decomposes over samples and causes no bias.
Thus we focus on the negative entropy term
\begin{align*}
- \wh{H}(Z) =  \sum_{z} \hat{q}(z) \log \hat{q}(z)
\end{align*}
whose gradient with respect to $q$ is
\begin{align}
  &\sum_{z} \paren{1 + \log \hat{q}(z)} \nabla \hat{q}(z) \notag \\
  &=  \frac{1}{K} \sum_{k=1}^K \sum_{z} \paren{1 + \log \hat{q}(z)} \nabla \hat{q}_k(z) \label{eq:a1}
\end{align}
where we expand $\nabla \hat{q}(z)$ by the identity
\begin{align}
  \nabla \hat{q}(z) = \frac{1}{N} \sum_{i=1}^N  \nabla q(z|y_i) = \frac{1}{K} \sum_{k=1}^K \nabla \hat{q}_k(z)    \label{eq:afact}
\end{align}
In contrast, the gradient of the negative entropy term averaged over minibatches is
\begin{align}
 \frac{1}{K} \sum_{k=1}^K \sum_{z} \paren{1 + \log \hat{q}_k(z)} \nabla \hat{q}_k(z) \label{eq:a2}
\end{align}
Hence the difference between \eqref{eq:a1} and \eqref{eq:a2} is
\begin{align*}
  \frac{1}{K} \sum_{k=1}^K \sum_{z} \log \frac{\hat{q}(z)}{\hat{q}_k(z)} \nabla \hat{q}_k(z)
\end{align*}
This shows the first result.
Now we analyze the generalized Brown loss
\begin{align*}
  \frac{1}{N} \sum_{i=1}^N \sum_{z, z'} p(z|x_i) q(z'|y_i) \log \frac{\hat{p}(z)\hat{q}(z')}{p(z|x_i) q(z'|y_i)}
\end{align*}
When we expand the log fraction, we see that the denominator decomposes over samples and causes no bias.
Thus we focus on the numerator term
\begin{align*}
  \frac{1}{N} \sum_{z, z'} \log \paren{\hat{p}(z)\hat{q}(z')} \sum_{i=1}^N  p(z|x_i) q(z'|y_i)
\end{align*}
By the product rule, its gradient with respect to $q$ is a sum of two terms.
The first term is (using \eqref{eq:afact} again)
\begin{align}
  \frac{1}{N}  \sum_{k=1}^K \sum_{z, z'} \paren{\frac{1}{K} \sum_{i=1}^N   \frac{p(z|x_i) q(z'|y_i)}{\hat{q}(z')}}  \nabla \hat{q}_k(z') \label{eq:a3}
\end{align}
The second term is (as a sum over batches)
\begin{align}
  \frac{1}{N} \sum_{k=1}^K  \sum_{z, z'}  \log \paren{\hat{p}(z)\hat{q}(z')}    \sum_{(x, y) \in B_k} p(z|x)  \nabla q(z'|y) \label{eq:a4}
\end{align}
In contrast, the numerator term estimated as an average over minibatches is
\begin{align*}
  \frac{1}{N}  \sum_{k=1}^K \sum_{z, z'} \log \paren{\hat{p}_k(z)\hat{q}_k(z')} \sum_{(x, y) \in B_k} p(z|x) q(z'|y)
\end{align*}
and the two terms of its gradient with respect to $q$ (corresponding to \eqref{eq:a3} and \eqref{eq:a4}) are
\begin{align}
  &\frac{1}{N} \sum_{k=1}^K   \sum_{z, z'} \paren{\sum_{(x, y) \in B_k}   \frac{p(z|x) q(z'|y)}{\hat{q}_k(z')}} \nabla \hat{q}_k(z') \label{eq:a5} \\
  &\frac{1}{N} \sum_{k=1}^K \sum_{z, z'} \log \paren{\hat{p}_k(z)\hat{q}_k(z')}   \sum_{(x, y) \in B_k}  p(z|x_i)  \nabla q(z'|y) \label{eq:a6}
\end{align}
Thus the difference between \eqref{eq:a3} and \eqref{eq:a5} is
\begin{align*}
  \frac{1}{N}\sum_{k=1}^K \sum_{z, z'} \ep_k(z,z') \nabla \hat{q}_k(z')
\end{align*}
where
\begin{align*}
  \ep_k(z,z') = &\frac{1}{K} \sum_{i=1}^N \frac{p(z|x_i) q(z'|y_i)}{\hat{q}(z')}   \\
  &- \sum_{(x, y) \in B_k} \frac{p(z|x) q(z'|y)}{\hat{q}_k(z')}
\end{align*}
The difference between \eqref{eq:a4} and \eqref{eq:a6} is
\begin{align*}
  \frac{1}{N}  \sum_{k=1}^K \sum_{z, z'}  \log \frac{\hat{p}(z)\hat{q}(z')}{\hat{p}_k(z)\hat{q}_k(z')}  \sum_{(x, y) \in B_k} p(z|x) \nabla q(z'|y)
\end{align*}
Adding these differences gives the second result.

\end{document}